\documentclass[a4paper]{article}

\usepackage{paper,math}
\usepackage[margin=1in]{geometry}
\usepackage{authblk}
\usepackage{tcolorbox}
\tcbuselibrary{breakable}

\newtcolorbox{promptbox}{
    breakable,
    colback=gray!5,
    colframe=black,
    boxrule=1pt,
    top=\baselineskip,
    bottom=\baselineskip,
    right=10pt,
    left=10pt,
}

\graphicspath{{figs/}}

\begin{document}

\title{Think in Blocks: Adaptive Reasoning from Direct Response to Deep Reasoning}

\author[1]{Yekun Zhu}
\author[2]{Guang Chen}
\author[3]{Chengjun Mao}
\affil[1]{{\small zhuyekun123@sjtu.edu.cn}}
\affil[2]{{\small cg234573@antgroup.com}}
\affil[3]{{\small chengjun.mcj@antgroup.com}}
\renewcommand{\thefootnote}{*}
\maketitle

\begin{abstract}
  Large Language Models (LLMs) with chains-of-thought have demonstrated strong performance on an increasing range of tasks, particularly those involving complex logical reasoning. However, excessively long chains can lead to overthinking, causing computational waste and slower responses. This raises a question: can LLMs dynamically adjust the length of their reasoning processes based on task complexity? To address this, we propose the Think in Blocks framework, which enables adaptive reasoning—from zero to deep reasoning—by partitioning the reasoning process into a tunable number of blocks. Our main contributions are: (1) Establishing an explicit block-structured paradigm in which the model first predicts an integer reasoning budget—the number of blocks—and then partitions its reasoning accordingly; (2) Training an adaptive model through a three-stage pipeline—Supervised Fine-Tuning, reward-guided Direct Preference Optimization, and Reinforcement Learning—that adjusts its reasoning depth to problem difficulty; (3) Exploiting the explicit block count to dynamically control reasoning depth at inference time, allowing flexible adjustment of chain-of-thought length during deployment.
\end{abstract}

\section{Introduction}
\label{sec:introduction}

Recently, Large Language Models (LLMs) have demonstrated remarkable capabilities across various tasks, particularly those requiring complex reasoning~\citep{touvron2023llama, anthropic2023claude, achiam2023gpt, yang2025qwen3}.
Their success in complex reasoning is mainly due to the use of chain-of-thought (CoT) prompting~\citep{wei2022chain}, which allows models to decompose multi-step problems into intermediate steps.

Despite this success, many have observed that large reasoning models exhibit a significant overthinking problem~\citep{chen2024not, feng2025efficient, shen2025dast}.
Specifically, they tend to expend excessive computational resources (in tokens or reasoning steps) on questions that are exceptionally simple or where the answer is self-evident.
The model tends to use roughly the same number of tokens for both simple and complex problems, revealing its inability to distinguish difficulty levels and adapt reasoning depth.

The overthinking problem in LLMs poses a significant challenge. To address this, researchers have explored several promising strategies.
These include using reinforcement learning to encourage shorter reasoning paths~\citep{team2025kimi, yu2025dapo, luo2025o1, fang2025thinkless}, applying methods like Direct Preference Optimization (DPO) to generate more concise chains of thought~\citep{shen2025dast}, merging models specialized in both extended and brief reasoning~\citep{ilharco2022editing, yadav2023ties, ma2025cot}, and developing hybrid approaches that allow models to dynamically switch between thinking and non-thinking modes based on the problem's demands~\citep{zhang2025adaptthink, lou2025adacot}.
However, these approaches face limitations. Most either apply a uniform reasoning process to all problems, failing to prevent overthinking on simple tasks, or only offer a binary choice: to think or not to think. A truly adaptive model should instead be able to dynamically adjust the depth of its reasoning.

\begin{figure}
  \begin{center}
    \includegraphics[width=0.85\textwidth]{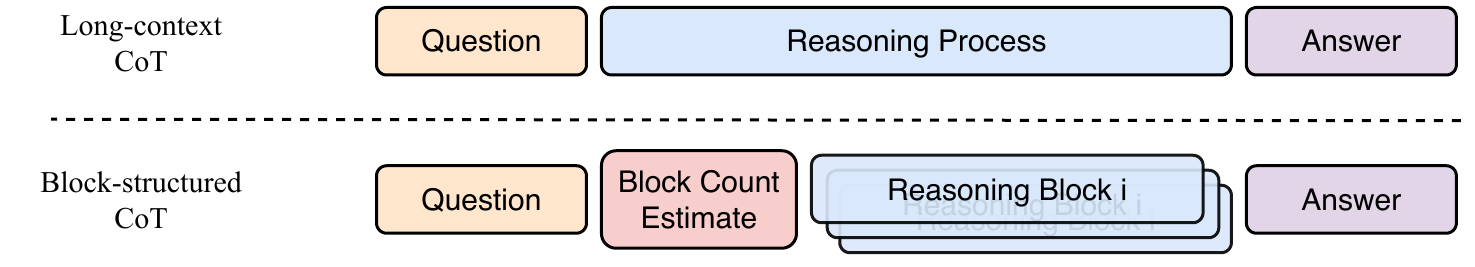}
  \end{center}
  \caption{Block-structured chain-of-thought (CoT). The model (1) reads the question and estimates the number of reasoning blocks; (2) completes each block with a coherent chain-of-thought or a full tool-call result; and (3) produces the final answer conditioned on these blocks.}
  \label{fig:overthinking_example}
\end{figure}

Motivated by these challenges, we propose the Think in Blocks framework to enable adaptive reasoning in LLMs.
Our approach partitions the reasoning process into discrete blocks, allowing for dynamic control over reasoning processes by adjusting the number of blocks. This enables the model to perform reasoning from zero to multiple steps as needed. Specifically, the model is trained to first assess the problem's complexity to obtain a corresponding number of reasoning blocks—each representing a distinct step or tool use—and then generate the content for each block.
To achieve this, we first construct a cold-start dataset to help the model predict the required number of blocks for a given problem. This dataset also exposes the model to examples of both non-thinking responses and deep-thinking responses. We then fine-tune the model using DAST-like Direct Preference Optimization and Reinforcement Learning (RL) with rewards based on mathematical optimization problems.

Our main contributions are as follows:
\begin{enumerate}
  \item We propose an explicit block-structured reasoning scheme that allows the model to estimate its reasoning budget and articulate the solution using that many blocks, thereby automatically matching reasoning depth to problem difficulty.
  \item We learn this format through a three-stage pipeline—supervised fine-tuning, reward-guided Direct Preference Optimization, and reinforcement learning—which teaches the model to shorten its reasoning for simple questions and elaborate for difficult ones.
  \item The explicit block count serves as a practical control at inference time, enabling users to cap or extend the number of blocks to trade off efficiency against accuracy as needed.
\end{enumerate}

\section{Related Works}
\label{sec:related_works}

\paragraph{Large Language Models.}
Recent Large Language Models (LLMs), such as OpenAI's o1~\citep{openai2024learning}, DeepSeek's DeepSeek-R1~\citep{guo2025deepseek} and Qwen3~\citep{yang2025qwen3} have developed the ability to perform human-like reasoning by generating long chains of thought before arriving at an answer. However, these models often struggle with overthinking, expending excessive computational resources on simple problems~\citep{chen2024not, feng2025efficient, shen2025dast, shojaee2025illusion}. This overthinking problem has been a significant focus of recent research.

\paragraph{Efficient Reasoning Models.}
To address the overthinking problem in LLMs, recent research has explored strategies to reduce reasoning length. Some approaches employ test-time methods like Prompt Engineering~\citep{chen2024not, feng2025efficient} to encourage models to generate shorter reasoning paths. Alternatively, RL with length penalty can optimize the reasoning process~\citep{team2025kimi, yu2025dapo, fang2025thinkless, yeo2025demystifying}. Other methods finetune the model using length-based preference pairs, typically obtained via best-of-N sampling~\citep{shen2025dast}. Additionally, some approaches merge models specialized in short CoT and long CoT~\citep{ilharco2022editing, yadav2023ties, ma2025cot} without requiring additional training.
However, these methods typically apply a uniform reasoning process to all problems. Crucially, non-thinking mode can sometimes outperform thinking mode, particularly for simple problems.

\paragraph{Hybrid Reasoning Models.}
Other researchers have proposed hybrid reasoning models that can dynamically switch between thinking and non-thinking modes based on the problem's complexity. This approach can be realized either through a system involving multiple models or within a single model. In multi-model frameworks, routing mechanisms select the appropriate model based on problem complexity~\citep{han2024token, chuang2024learning, ong2024routellm}. For example, a lightweight model may generate an answer and verify whether it is correct; if unconfident, it will call a more powerful model to generate a longer reasoning chain.
In contrast, single-model frameworks train the model to predict the required reasoning depth and adjust its response accordingly~\citep{zhang2025adaptthink, lou2025adacot}. However, these approaches typically offer only a binary choice: to think or not to think. Critically, a truly adaptive model should be capable of dynamically adjusting its reasoning depth.
\section{Block-structured Chain-of-Thought}
\label{sec:block_structured_cot}

The core idea is to partition the reasoning process into a variable number of \textbf{reasoning blocks}, then adjust this number to enable zero, fast, or deep thinking. Concretely, the model is trained to: (1) read the question and, based on an initial impression, output the total budget for reasoning blocks; (2) complete each reasoning block by generating a coherent chain of thought or a complete tool-call result; and (3) output the final response based on the reasoning blocks.

We introduce the block-count estimator for two reasons:
\begin{enumerate}
    \item During training (DPO and RL phases) it serves as an optimizable signal: the model learns to allocate more blocks to hard questions and fewer to easy ones, thereby improving overall performance.
    \item At inference time it serves as an explicit parameter for setting the reasoning budget. When computing resources or latency budgets are tight, the user can lower the block cap to trade a little accuracy for faster responses; when resources are ample, the cap can be raised—or removed—to pursue higher accuracy. Such fine-grained control over reasoning depth is hard to achieve with the standard CoT approach.
\end{enumerate}

Specifically, Table~\ref{tab:cot} shows the block-structured formats of reasoning process that we want the model to generate.

\begin{table}[H]
    \begin{center}\caption{Prompt structure comparison of Long-context CoT (a single continuous reasoning block) and Block-structured CoT (\tttag{thought\_segments}-scaled reasoning blocks separated by \tttag{continue\_think} markers).}\label{tab:cot}
        \begin{tabular}{@{}l|l@{}}
            \toprule
            $\textbf{Long-context CoT}$                                                                                                                                     & $\textbf{Block-structured CoT}$                                                                                                                                                                                                                                                                                                                                                                                                                                                                                          \\ \midrule
            \begin{tabular}[c]{@{}l@{}}\textless{}think\textgreater\\ \\ \\ \{reasoning process\}\\ \\ \\ \\ \\ \textless{}/think\textgreater\\ {\{response\}}\end{tabular} & \begin{tabular}[c]{@{}l@{}}\textless{}think\textgreater\\     \tagblock{red!70}{\textless{}thought\_segments\textgreater{}n\textless{}/thought\_segments\textgreater}\\     \{reasoning process 1\}\\     \tagblock{blue!50}{\textless{}continue\_think\textgreater}\\     \{reasoning process 2\}\\     ...\\     \tagblock{blue!50}{\textless{}continue\_think\textgreater}\\     \{reasoning process n\}\\ \textless{}/think\textgreater\\ {\{response\}}\end{tabular} \\ \bottomrule
        \end{tabular}
    \end{center}
\end{table}

\section{Method}
\label{sec:method}
The Think in Blocks framework is implemented in three stages: (1) \textbf{Supervised Fine-Tuning}, where we train the model to learn the paradigm for predicting reasoning budgets
and partitioning the reasoning process into blocks, as well as exposes the model to examples of both non-thinking responses and deep-thinking responses; (2) \textbf{Direct Preference Optimization}, where we fine-tune the model using DAST-like Direct Preference Optimization; and (3) \textbf{Reinforcement Learning}, where we formulate the optimization process as an objective function within RL rewards.

\subsection{Supervised Fine-Tuning}
Supervised Fine-Tuning here is mainly used to teach the model to generate the desired format of reasoning process.

\subsubsection{Dataset Construction}

Given a prompt set $\mathcal{X} = \{x_i\}_{i=1}^{N}$, we use LLMs to generate the reasoning process and response $\{y_i\}_{i=1}^{N}$.
Then, we use prompt engineering to refactor $\{y_i\}_{i=1}^{N}$ and split the reasoning process into multiple blocks, donate the refactored $\{y_i\}_{i=1}^{N}$ with block-structured format as $\{\haty_i\}_{i=1}^{N}$:
\begin{equation}
    \mathcal{D}_{\text{SFT}} = \{(x_i, \hat{y}_i)\}_{i=1}^{N}.
\end{equation}
Here, each refactored response $\hat{y}_i$ is structured as a sequence of $n_{\hat{y}_i}$ reasoning blocks $(B^{\hat{y}_i}_1,\dots,B^{\hat{y}_i}_{n_{\hat{y}_i}})$ followed by the final solution. The integer $n_{\hat{y}_i}$ comes from the \texttt{\textless{}thought\_segments\textgreater} tag.\label{subsubsec:response-structure}

\begin{figure}[H]
    \begin{center}
        \includegraphics[width=0.7\textwidth]{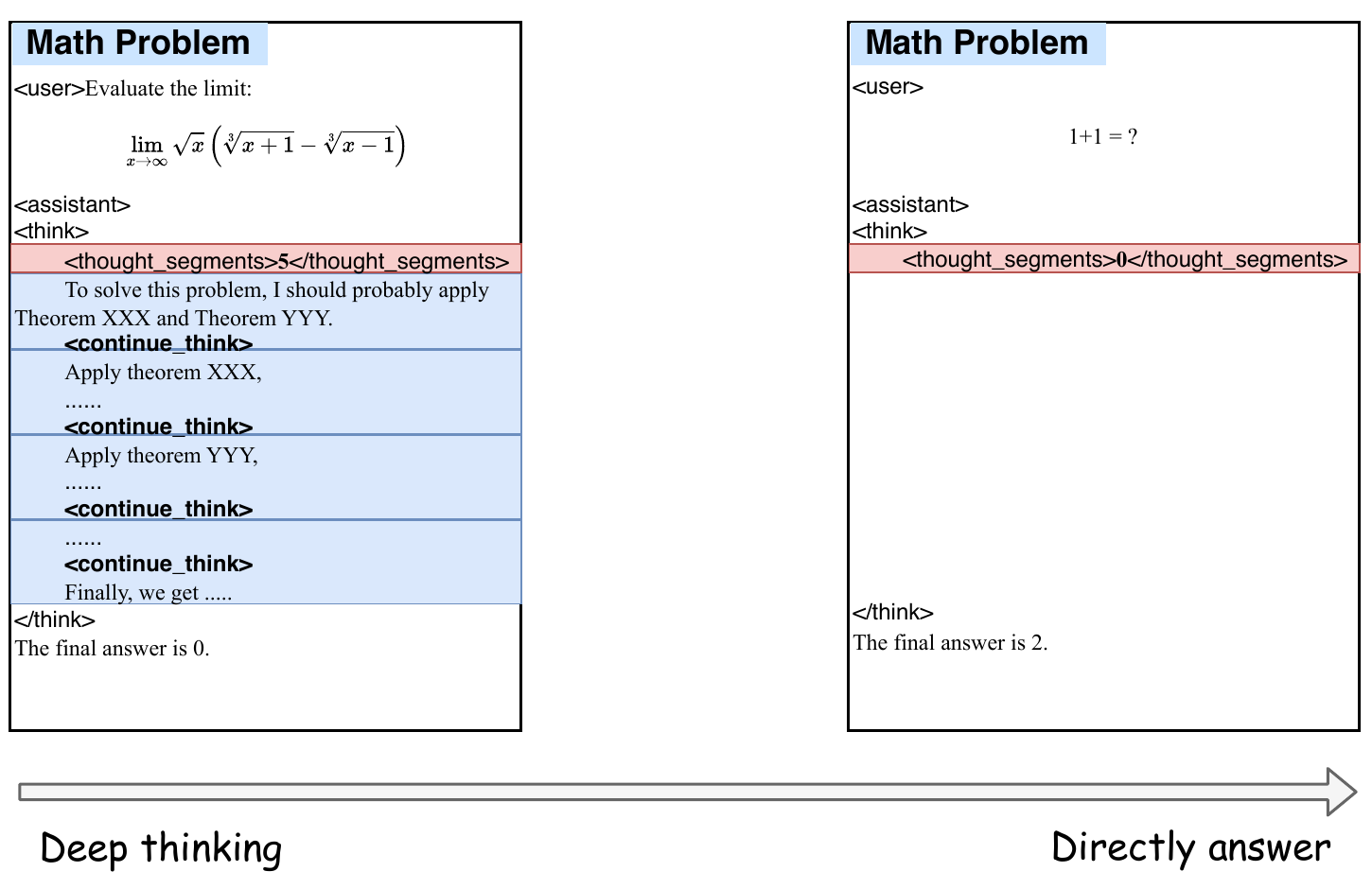}
    \end{center}
    \caption{Illustration of block-structured reasoning. Left example: deep thinking with multiple reasoning blocks; right example: no thinking where the model predicts the answer directly. The arrow highlights the adaptive transition from detailed reasoning to a direct response as problem difficulty decreases.}
\end{figure}

However, two edge cases require special handling during dataset construction:
(1) \textbf{Non-Thinking}: For trivial problems we remove the reasoning trace entirely and set the number of reasoning blocks to~0.
(2) \textbf{Deep-Thinking}: For short yet challenging problems we expand the chain of thought into additional reasoning blocks through prompt engineering and knowledge distillation.

\subsubsection{Multi-stage Training}

Inspired by curriculum learning, we divide the supervised fine-tuning into two stages. In the first stage, we train the model on medium-difficulty examples containing a moderate number of reasoning blocks. This stage focuses exclusively on learning the correct output format, and the two special cases mentioned earlier (Non-Thinking and Deep-Thinking instances) are excluded from this stage. The reason is that we aim to avoid a negative impact on reasoning ability: data lacking reasoning steps might affect accuracy, while long reasoning chains could lead to overthinking.

In the second stage, we add a small number of both Non-Thinking and Deep-Thinking examples to the training data. However, most of the data still consists of medium-difficulty prompts.
This approach gives the model limited exposure to the full range of reasoning complexities—from simple to advanced. The key goal is to prevent these special cases from overwhelming what the model learns.

\subsection{Direct Preference Optimization}

In this stage, we fine-tune the policy with Direct Preference Optimization (DPO). Our goal is to convert the high-level requirement of adaptive reasoning---using fewer blocks for simple problems and more blocks for difficult ones---into a scalar reward that can automatically guide pair construction. This reward scores each sampled response by jointly assessing its accuracy and whether its overall length is appropriate for the prompt's difficulty.

Consequently, given multiple responses for the same prompt, we rank them with this reward and label the higher-reward response as chosen and the lower-reward response as rejected whenever their reward gap exceeds a threshold. These chosen-rejected pairs provide the preference signal that DPO optimizes against.

To construct such pairs, we first sample $s$ responses for each of $N$ prompts from the current policy $\pi_\theta$, denoted as $\mathcal{D}_{\text{DPO}}=\{(x_i, \hat{y}_i^{(1)}, \cdots, \hat{y}_i^{(s)})\}_{i=1}^{N}$, where $\hat{y}_i^{(j)} \sim \pi_\theta(\cdot|x_i)$ for $j=1,\dots,s$. Note that these sampled responses are not necessarily correct, and their rewards will be calibrated in the next subsection.

\subsubsection{DAST Reward Score Calibration}
We use a Direct Preference Optimization approach inspired by Difficulty-Adaptive Slow Thinking (DAST) to fine-tune the model~\citep{shen2025dast}.
DAST quantifies response quality by favoring shorter responses for easier problems and longer responses for harder problems.

DAST dynamically adjusts the reasoning process based on problem difficulty, with its core mechanism establishing a relationship between response length and difficulty. This relationship enables the construction of pairwise datasets for DPO training.

However, since standard DAST applies a uniform reasoning process to all problems, we extend it into a more flexible framework. Our approach allows the LLM to adapt its reasoning process from direct responses to deep thinking by strategically incorporating both direct-answer and deep-thinking instances when constructing chosen and rejected pairs for DPO.

The DAST's Token Length Budget (TLB) metric is formally defined as:
\begin{equation}
    L_{\text{budget}}(x_i) = p\cdot L_{\bar{r}}(x_i) + L_{\text{max}}(x_i),
\end{equation}
where $p = N_\text{cor} / {s}$ denotes the ratio of correct respones to the total number of sampling respones, and $L_{\bar{r}}, L_{\text{max}}$ is the expectation of the generation length and maximum generation length, respectively. Specifically,
\begin{equation}
    L_{\bar{r}}(x_i) = \mathbb{E}_{y\sim \pi_\theta(\cdot|x_i)}[L(y)]  \approx \frac{1}{s}\sum_{j=1}^{s} L(\hat{y}_i^{(j)}),
\end{equation}
where $L(y)$ is the length of the response $y$.

Then we can compute the reward score calibration as:
\begin{equation}
    \begin{aligned}
        r(\hat{y}_i^{(j)}) = \begin{cases}
                                 \max(-0.5\lambda(\hat{y}_i^{(j)}) + 0.5, 0.1) & \text{if correct} \\
                                 \min(0.9 \lambda(\hat{y}_i^{(j)}) -0.1, -0.1) & \text{otherwise}
                             \end{cases},
    \end{aligned}
\end{equation}
where $\lambda(\hat{y}_i^{(j)}) = \frac{L(\hat{y}_i^{(j)}) - L_{\text{budget}}(x_i)}{L_{\text{budget}}(x_i)}$.

With the calibrated reward scores and a threshold $\delta$, we form chosen-rejected pairs from the sampled responses: specifically, for any two responses $\hat{y}_i^{(j)}$ and $\hat{y}_i^{(k)}$, we designate the higher-reward one as chosen and the lower as rejected only if $r(\hat{y}_i^{(j)}) - r(\hat{y}_i^{(k)}) > \delta$ and the number of reasoning blocks in the chosen response is no greater than in the rejected one.

\subsection{Reinforcement Learning}
Although SFT teaches the model to produce block-structured reasoning formats and DPO encourages initial adaptation to problem difficulty, reinforcement learning further enhances this capability by optimizing the balance between reasoning depth (via number of reasoning blocks) and overall efficiency through a tailored reward function.

Let $r$ be a reward model of the proposed answer $\hat{y}$ for the given problem $x$ based on the ground truth $\hat{y}^*$. Reviewing our requirements, we design the following reward function to guide the model:

\begin{takeawaybox}{Objectives}
    \begin{itemize}\setlength{\itemsep}{0pt}\setlength{\parskip}{0pt}\setlength{\parsep}{0pt}\setlength{\topsep}{0pt}
        \item Maximize No-Thinking samples
        \item Minimize reasoning block count
        \item Reduce average reasoning block length
        \item Ensure format consistency
        \item Maintain/improve task accuracy
    \end{itemize}
\end{takeawaybox}

To formalize this, recalling the response structure from Subsection~\ref{subsubsec:response-structure}, the RL optimization aims to maximize the probability of No-Thinking responses (i.e., $n_{\hat{y}}=0$) while minimizing the expected number of blocks, the average block length, and ensure format consistency, relative to a reference policy $\pi_{\theta_\text{ref}}$, subject to maintaining or improving the expected task reward $r(x,\hat{y},\hat{y}^*)$:
\begin{align*}
     & \max_\theta \mathbb{E}_{x\sim\mathcal{D}, \hat{y}\sim\pi_\theta(\cdot|x)} [\mathbbm{1}\{n_{\hat{y}} = 0\}]                                                                                            \\
     & \min_\theta \mathbb{E}_{x\sim\mathcal{D},\hat{y}\sim\pi_\theta(\cdot|x)} [n_{\hat{y}}]                                                                                                                \\
     & \min_\theta \mathbb{E}_{x\sim\mathcal{D},\hat{y}\sim\pi_\theta(\cdot|x)} \left[ \frac{1}{n_{\hat{y}}} \sum_{i=1}^{n_{\hat{y}}} |B^{\hat{y}}_i| \right]                                                \\
     & \min_\theta \mathbb{E}_{x\sim\mathcal{D},\hat{y}\sim\pi_\theta(\cdot|x)} \left[ |n_{\hat{y}}-\tilde{n}_{\hat{y}}| \right]                                                                             \\
     & \text{s.t.} \quad \mathbb{E}_{x\sim\mathcal{D},\hat{y}\sim\pi_\theta(\cdot|x)} [R(x,\hat{y}, \hat{y}^*)] > \mathbb{E}_{x\sim\mathcal{D},y'\sim\pi_{\theta_\text{ref}}(\cdot|x)} [R(x,y', \hat{y}^*)],
\end{align*}
where $R(x,\hat{y}, \hat{y}^*)$ is a rule-based correctness reward that returns 1 when the proposed answer is correct and 0 otherwise. We denote by $n_{\hat{y}}$ the block count declared in the \texttt{<thought\_segments>} tag and by $\tilde n_{\hat{y}}$ the actual number of reasoning blocks produced; the term $|n_{\hat{y}}-\tilde n_{\hat{y}}|$ thus enforces format consistency by penalising any mismatch. Finally, $\mathbbm{1}\{n_{\hat{y}} = 0\}$ is the indicator function of the no-thinking mode.

We introduce non-negative Lagrange multipliers $\boldsymbol{\lambda}=(\lambda_1,\lambda_2,\lambda_3,\lambda_4)$ and move the constraints into the objective. The resulting Lagrangian is
\begin{equation}\label{eq:lagrangian}
    \begin{aligned}
        \mathcal{L}(\theta, \boldsymbol{\lambda})={} & \Bigl(\mathbb{E}_{x,\hat y\sim\pi_{\theta}}[R(x,\hat y,\hat y^*)]-\mathbb{E}_{x,y'\sim\pi_{\theta_{\text{ref}}}}[R(x,y',\hat y^*)]\Bigr) \\[4pt]
                                                     & + \lambda_1\,\mathbb{E}_{x,\hat y\sim\pi_{\theta}}\bigl[\mathbbm{1}\{n_{\hat y}=0\}\bigr]                                                \\[2pt]
                                                     & - \lambda_2\,\mathbb{E}_{x,\hat y\sim\pi_{\theta}}\bigl[n_{\hat y}\bigr]                                                                 \\[2pt]
                                                     & - \lambda_3\,\mathbb{E}_{x,\hat y\sim\pi_{\theta}}\Bigl[\tfrac{1}{n_{\hat y}}\sum_{i=1}^{n_{\hat y}} |B^{\hat y}_i|\Bigr]                \\[2pt]
                                                     & - \lambda_4\,\mathbb{E}_{x,\hat y\sim\pi_{\theta}}\bigl[|n_{\hat y}-\tilde n_{\hat y}|\bigr].
    \end{aligned}
\end{equation}

The reference term $\mathbb{E}_{x,y'\sim\pi_{\theta_{\text{ref}}}}[R(x,y',\hat y^*)]$ is approximated with $s$ Monte-Carlo samples
\begin{equation}\label{eq:reward-estimate}
    \hat R_{\text{ref}}(x)= \frac{1}{s}\sum_{j=1}^{s} R\bigl(x,y'_j,\hat y^*\bigr).
\end{equation}

\paragraph{Advantage.} Defining the per-sample advantage
\begin{equation}\label{eq:advantage}
    A\bigl(x,\hat y,\hat y^*\bigr)= R\bigl(x,\hat y,\hat y^*\bigr)-\hat R_{\text{ref}}(x)
    +\lambda_1\,\mathbbm{1}\{n_{\hat y}=0\}-\lambda_2\,n_{\hat y}-\lambda_3\,\frac{1}{n_{\hat y}}\sum_{i=1}^{n_{\hat y}}|B^{\hat y}_i| - \lambda_4\,|n_{\hat y}-\tilde n_{\hat y}|,
\end{equation}
we can optimise $\theta$ with a PPO-style surrogate objective
\begin{equation}\label{eq:ppo-loss}
    \mathcal{L}_{\text{PPO}}(\theta)=\mathbb{E}_{x,\hat y\sim\pi_{\theta_{\text{old}}}}\Bigl[\min\bigl(r_{\theta}(x,\hat y)\,A,\;\text{clip}\bigl(r_{\theta}(x,\hat y),1-\epsilon,1+\epsilon\bigr)\,A\bigr)\Bigr],
\end{equation}
where $r_{\theta}(x,\hat y)=\tfrac{\pi_{\theta}(\hat y|x)}{\pi_{\theta_{\text{old}}}(\hat y|x)}$.

\paragraph{Implementation notes.}  However, in practice, because there are too many hyperparameters that directly affect the final objective in advantage function~\ref{eq:advantage}, the PPO-style loss is too complex and not stable. To address this, we adapt the multipliers through accuracy-aware scaling. The coefficients $\lambda_i$ control the trade-off between accuracy and reasoning cost. In practice we either (i) tune them manually, or (ii) adapt them with a factor such as accuracy-dependent scaling. After each rollout we compute the empirical accuracy $p$ and derive a scaling factor
\begin{equation}\label{eq:h_scale}
    h(p)=\operatorname{clip}\left(\frac{p-p_{\text{low}}}{p_{\text{high}}-p_{\text{low}}},\;0,\;1\right),
\end{equation}
which linearly interpolates between 0 and 1.  We then rescale the base hyper-parameters $\lambda_i^{\!*}$ as
\begin{equation}\label{eq:lambda_scale}
    \lambda_i = h(p)\,\lambda_i^{\!*}, \qquad i\in\{1,2,3,4\}.
\end{equation}
When accuracy is low (small $p$) the model focuses on improving correctness; as accuracy rises, the penalty on reasoning length becomes more significant.


Table~\ref{tab:rl_coeffs} summarises the coefficients and thresholds used in the RL reward function.
\begin{table}[H]
    \begin{center}
        \begin{tabular}{@{}lcc@{}}
            \toprule
            \textbf{Name}                      & \textbf{Symbol}   & \textbf{Description}                                                           \\\midrule
            \texttt{nothink\_bonus\_coef}      & $\lambda_{1}^*$   & bonus for no-thinking mode $\mathbbm{1}\{n_{\hat{y}}=0\}$                      \\
            \texttt{count\_coef}               & $\lambda_{2}^*$   & penalty on reasoning block count $n_{\hat{y}}$                                 \\
            \texttt{block\_len\_coef}          & $\lambda_{3}^*$   & penalty on average block length $\tfrac{1}{n_{\hat{y}}}\sum_i |B^{\hat{y}}_i|$ \\
            \texttt{seg\_count\_coef}          & $\lambda_{4}^*$   & penalty on format mismatch $|n_{\hat{y}}-\tilde{n}_{\hat{y}}|$                 \\ \midrule
            \texttt{accuracy\_threshold\_low}  & $p_{\text{low}}$  & lower bound of accuracy-aware scaling                                          \\
            \texttt{accuracy\_threshold\_high} & $p_{\text{high}}$ & upper bound of accuracy-aware scaling                                          \\\bottomrule
        \end{tabular}
        \caption{Reward coefficients and thresholds used during RL fine-tuning.}\label{tab:rl_coeffs}
    \end{center}
\end{table}

\section{Inference-Time Control of Reasoning Block Count}
\label{sec:test_time}

In the training phase, making the block count explicit gives DPO and RL a concrete signal to optimise: the model learns to produce short traces for easy questions while allocating more blocks to hard ones. At inference time, the same prediction turns into a tunable reasoning budget. When latency or energy are limited—for example on edge devices—the user can cap the number of blocks to obtain faster answers with only a minor loss in accuracy; conversely, in accuracy-critical scenarios the cap can be relaxed or removed altogether to allow deeper reasoning and higher performance.

Based on this, we support two inference modes:
\begin{enumerate}
    \item \textbf{Auto mode} — the model uses its own block-count prediction.
    \item \textbf{Override mode} — the caller specifies an upper (or lower) bound for the number of blocks, and the model must reason within that range.
\end{enumerate}

\begin{algorithm}[H]
    \caption{Override-BlockCap Decoding}\label{alg:block-cap-decoding}
    \vspace{0.4em}
    $\Def\; \text{BlockCapDecode}\bigl(\texttt{prompt},\, \texttt{cap\_low},\, \texttt{cap\_high}\bigr)$
    \begin{algorithmic}[1]
        \vspace{0.4em}
        \Comment{\textit{Step 1: predict logits for block count $k$}}
        \State $\boldsymbol \ell \gets \text{PredictLogits}(\texttt{prompt})$ \hfill $\boldsymbol \ell\in\mathbb R^{K}$
        \vspace{0.4em}
        \Comment{\textit{Step 2: mask logits outside user range}}
        \ForAll{$i$}\; $\boldsymbol\ell_i \gets -\infty$ \textbf{if} $i\notin[\texttt{cap\_low},\texttt{cap\_high}]$ \EndFor
        \vspace{0.4em}
        \Comment{\textit{Step 3: sample $k$ from masked distribution}}
        \State $k \gets \text{SampleSoftmax}(\boldsymbol \ell)$ \hfill $k\in\mathbb N$
        \vspace{0.4em}
        \Comment{\textit{Step 4: generate response conditioned on $k$}}
        \State \texttt{response} $\gets \text{Generate}(\texttt{prompt},\, k)$
        \vspace{0.4em}
        \Ret $\texttt{response}$
    \end{algorithmic}
\end{algorithm}

The key idea is to zero out probabilities of block counts that violate the user constraint, renormalize, and continue decoding conditioned on the clamped value; the remainder of the generation process is unchanged. Alg.~\ref{alg:block-cap-decoding} illustrates this process.

The proposed block-cap mechanism offers practical advantages over the binary ``think/no-think'' control that simply switches the chain of thought on or off.
First, it provides fine-grained control: by adjusting the integer block budget~$k$ the user can change the depth of reasoning one block at a time, whereas the two-state gate offers only an all-or-nothing choice.
Second, it yields a smooth speed--accuracy trade-off: increasing~$k$ gradually improves accuracy while requiring more computation, so the user can choose any point between the two extremes.
Finally, the mechanism is architecture agnostic and requires no retraining. Because the cap is enforced by masking logits at decoding time, any model trained with block annotations can adopt it immediately.

\section{Experiments}
\label{sec:experiments}

\subsection{Setup}
\paragraph{LLMs and Datasets.}
We use Qwen3-8B~\citep{yang2025qwen3} as the base model.
To construct the training corpus, we start from the DeepMath~\citep{he2025deepmath} benchmark but retain only the problem statement and its annotated difficulty.
The reference answers are regenerated with Qwen3-32B because the original DeepSeek \texttt{think} chains-of-thought are overly long and differ stylistically from the Qwen3 family; mixing them causes a significant drop in performance.
For prompt engineering, we then run Qwen3 in the \texttt{no-think} mode to segment each regenerated answer into several reasoning blocks.
Consequently, each training example contains
(i)~\colorbox{red!15}{\texttt{\textless{}thought\_segments\textgreater}} specifying the number of blocks, and
(ii)~\colorbox{blue!15}{\texttt{\textless{}continue\_think\textgreater}} marking the boundary between consecutive blocks.
For high-difficulty problems, we further let Qwen3-32B perform self-reflection and rewrite its own answers to enrich the data.
More details of the implementation can be found in the Appendix~\ref{sec:appendix}.

\paragraph{Training Details.}
We train on $8\times$ A100 GPUs. In the SFT stage, we use a batch size of 8 and train for 3 epochs with learning rates of $8\times 10^{-5}$ (stage~1) and $2\times 10^{-6}$ (stage~2). The training set contains 7k samples, comprising 0.2k easy, 5k medium, and 1.8k hard problems. In the DPO stage, we use a batch size of 16, train for 3 epochs with a learning rate of $2\times 10^{-6}$, and construct 4k chosen--rejected pairs using the method mentioned above.
SFT and DPO are performed with the SWIFT framework~\citep{zhao2025swift}. In the RL stage, we experiment with three reward-based schemes---Group Relative Policy Optimization (GRPO)~\citep{shao2024deepseekmath}, its differentiable variant Dr.GRPO~\citep{liu2025understanding}, and Decoupled Clip and Dynamic sAmpling Policy Optimization (DAPO)~\citep{yu2025dapo}. Each method is trained for 2~epochs with a batch size of~32 and a learning rate of $1\times 10^{-6}$.
RL training is conducted with the verl framework~\citep{sheng2025hybridflow}.
The clipping ratios for DAPO are set to $\texttt{clip\_ratio\_low}=0.2$ and $\texttt{clip\_ratio\_high}=0.28$.
More details on RL training can be found in Section~\ref{sec:main_results}.

\begin{table}[H]
    \centering
    \begin{tabular}{@{}lccccc@{}}
        \toprule
        $\lambda_{1}^*$ & $\lambda_{2}^*$ & $\lambda_{3}^*$ & $\lambda_{4}^*$ & $p_{\text{low}}$ & $p_{\text{high}}$ \\
        \midrule
        0.1             & 0.05            & 0.5             & 0.1             & 0.75             & 0.9               \\
        \bottomrule
    \end{tabular}
    \caption{Hyperparameters for RL reward used in experiments.}
    \label{tab:rl_dapo_hparams}
\end{table}

\subsection{Main Results}
\subsubsection{Performance}

\label{sec:main_results}
\begin{table}[H]
    \begin{center}
        \begin{tabular}{@{}l|cc|c|ccc|cc@{}}
            \toprule
            \multirow{2}{*}{\textbf{Method}} & \multicolumn{2}{c|}{\textbf{Length}} & \multirow{2}{*}{\textbf{Bad cases (\%)}} & \multicolumn{3}{c|}{\textbf{Accuracy}} & \multirow{2}{*}{\textbf{$\Delta$Acc}} & \multirow{2}{*}{\textbf{$\Delta$Len(\%)}}                                                                        \\
                                             & mean                                 & std                                      &                                        & Overall                               & Easy                                      & Difficult &                            &                             \\
            \midrule
            Baseline                         & 7735                                 & 4395                                     & 2.2\%                                  & 85.5\%                                & 88.8\%                                    & 74.9\%    & --                         & --                          \\
            ~+GRPO                           & 7554                                 & 4957                                     & 1.0\%                                  & 85.9\%                                & 89.3\%                                    & 71.6\%    & +0.4\%                     & -2.3\%                      \\
            ~+$\text{GRPO}_\text{cosine}$    & 697                                  & 1365                                     & 0.6\%                                  & 58.8\%                                & 48.8\%                                    & 40.1\%    & -26.7\%                    & -91.0\%                     \\
            \midrule
            \multicolumn{7}{l}{\textit{Ours}}                                                                                                                                                                                                                                                                                      \\
            \midrule
            SFT-stage1                       & 5879                                 & 3132                                     & 19.7\%                                 & 67.3\%                                & 77.2\%                                    & 52.6\%    & -18.2\%                    & -24.0\%                     \\
            SFT-stage2                       & 5507                                 & 2853                                     & 27.8\%                                 & 61.6\%                                & 62.2\%                                    & 44.6\%    & \colorbox{red!15}{-23.9\%} & -28.8\%                     \\
            ~+DPO                            & 6077                                 & 4005                                     & 5.1\%                                  & 81.8\%                                & 82.9\%                                    & 72.5\%    & -3.7\%                     & \colorbox{red!15}{-21.5\%}  \\
            ~~~+DAPO                         & 5912                                 & 3837                                     & 5.1\%                                  & 83.1\%                                & 76.5\%                                    & 74.9\%    & -2.4\%                     & -23.6\%                     \\
            ~~~+Dr.GRPO                      & 3970                                 & 2373                                     & 1.5\%                                  & 77.2\%                                & 80.6\%                                    & 70.1\%    & -8.3\%                     & \colorbox{blue!15}{-48.7\%} \\
            ~~~+GRPO                         & 5791                                 & 3160                                     & 3.8\%                                  & 85.3\%                                & 92.0\%                                    & 74.9\%    & \colorbox{blue!15}{-0.2\%} & -25.1\%                     \\ \bottomrule
        \end{tabular}
        \caption{Results on the DeepMath~(1k) test set. Columns show mean answer length, its standard deviation, bad-case ratio, and accuracy on the overall/easy/difficult splits. $\Delta$Acc and $\Delta$Len report the change relative to the Baseline. Cells highlighted in \colorbox{blue!15}{blue} indicate desirable results (higher accuracy or greater length reduction), whereas cells in \colorbox{red!15}{red} indicate undesirable results (larger accuracy drop or smaller length reduction). The \emph{easy} split corresponds to difficulty levels 2.0--4.5, whereas the \emph{difficult} split corresponds to 8.0--9.0. Rows under \textit{Ours} use training data with block-structured chains-of-thought, and their DPO pair construction and RL reward follow the settings described in Section~\ref{sec:method}.}\label{tab:main_results}
    \end{center}
\end{table}

Table~\ref{tab:main_results} shows the test performance on a 1k problem subset that is randomly sampled from the DeepMath benchmark and is independent of the training data. The sampling preserves the original difficulty distribution, ensuring that the test set is i.i.d. with respect to the training dataset.
Length statistics (mean/std) are computed only for correct predictions; we ignore wrong answers because they always contain repetitive substrings that push generation to the maximum length limit.
We use the reward function introduced in~\citep{yeo2025demystifying} for $\text{GRPO}_\text{cosine}$.



We have the following observations:
\begin{enumerate}[label=(\arabic*)]
    \item Considering both metrics, the most effective pipeline is SFT $\rightarrow$ DPO $\rightarrow$ GRPO; it shortens answers by 25.1\% while keeping overall accuracy virtually unchanged ($\Delta\text{Acc}=-0.2\%$).
    \item The greatest reduction in length (--48.7\%) is achieved by Dr.GRPO, but at the cost of a considerable drop in accuracy ($-8.3\%$). Conversely, the largest accuracy loss ($-23.9\%$) occurs after the SFT2 stage, where we deliberately mix No-Think and Deep-Think instances; this is expected because switching reasoning modes hurts performance before RL can repair it.
    \item In practice, it is \emph{very hard} to simultaneously shorten answers substantially while also improving accuracy. Achieving both would require the model to radically alter its reasoning style and internal logic, which is costly and often unstable.
\end{enumerate}
\begin{figure}[H]
    \centering
    \begin{subfigure}[t]{0.48\linewidth}
        \centering
        \includegraphics[width=0.9\linewidth]{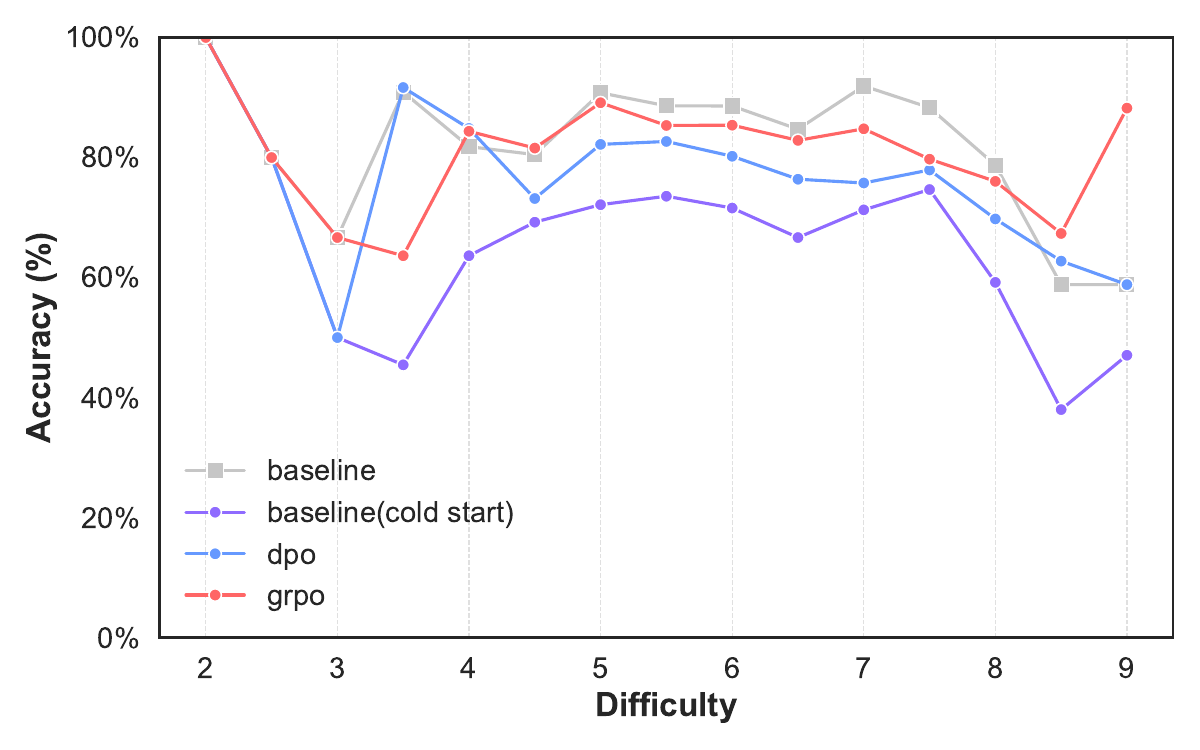}
        \caption{Accuracy by difficulty.}
    \end{subfigure}
    \hfill
    \begin{subfigure}[t]{0.48\linewidth}
        \centering
        \includegraphics[width=0.9\linewidth]{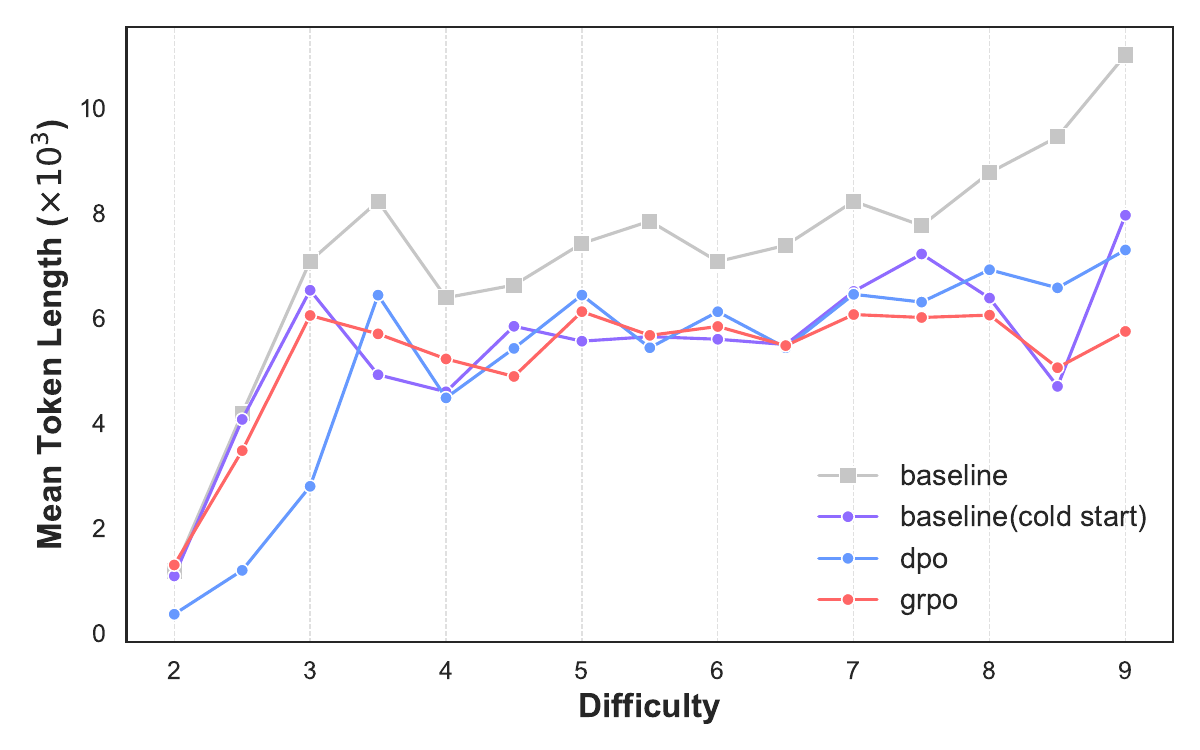}
        \caption{Length by difficulty.}
    \end{subfigure}
    \caption{Accuracy and average answer length across difficulty levels on the DeepMath~(1k) test set.}
    \label{fig:difficulty_curves}
\end{figure}

As shown in Fig.~\ref{fig:difficulty_curves}, our final model surpasses (or at least matches) the baseline on easy questions and on the hardest questions.
\subsubsection{Block Analysis}

\begin{figure}[H]
    \centering
    \begin{subfigure}[t]{0.48\linewidth}
        \centering
        \includegraphics[width=0.87\linewidth]{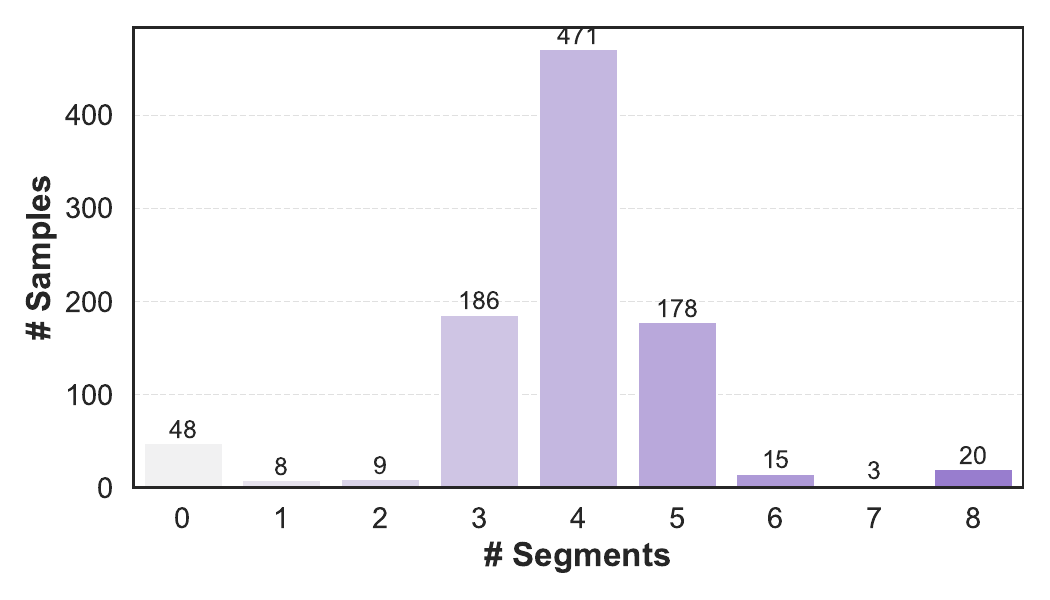}
        \caption{Number of samples by block count.}\label{fig:segments_vs_samples}
    \end{subfigure}
    \hfill
    \begin{subfigure}[t]{0.48\linewidth}
        \centering
        \includegraphics[width=\linewidth]{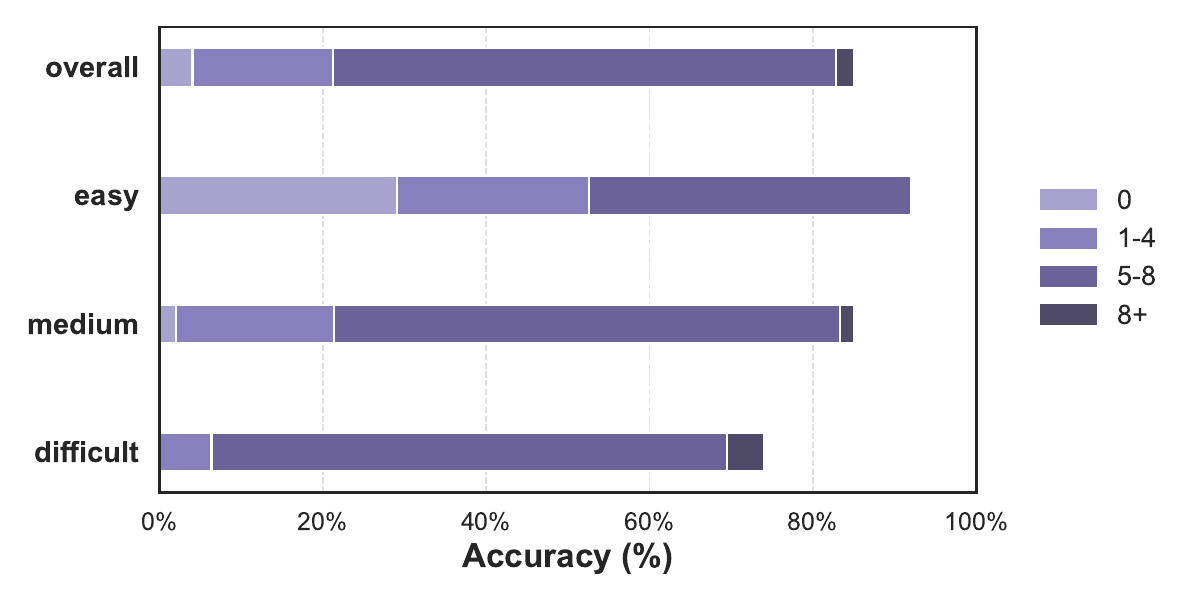}
        \caption{Accuracy by block count across difficulty levels.}\label{fig:accuracy_by_block_count}
    \end{subfigure}
    \caption{Block statistics and accuracy on the DeepMath~(1k) test set. Left: number of test samples for each block count (0 denotes no-think). Right: accuracy across difficulty levels; within each bar, colors indicate the proportion of samples for each block count.}
    \label{fig:block_analysis}
\end{figure}

Figure~\ref{fig:segments_vs_samples} shows the distribution of block counts in the test set. A block count of 0 denotes no-think responses. Samples with at least one block far outnumber those with zero blocks, indicating that deep-think responses are more common. Block counts of 3--5 are the most frequent, likely reflecting the difficulty distribution: medium-difficulty problems are more common than very easy or very hard ones.

Figure~\ref{fig:accuracy_by_block_count} displays accuracy by difficulty using stacked bars, where each bar is subdivided into colored segments that denote the percentage of predictions at each block count.
On easy problems (2.0--4.5), predictions are mostly produced with a block count of 0, indicating that the no-think mode is typically sufficient and deep-think ($\geq 8$ blocks) is rarely necessary. On medium-difficulty problems (4.5--7.5), the model most often uses 3--5 blocks. On hard problems (7.5--9.0), the no-think mode is seldom used and higher block counts are more common, reflecting the need for deeper reasoning.

\subsubsection{Inference-Time Control of Reasoning Block Count}

We use the method metioned in~\ref{sec:test_time} to adjust the block count at inference time. Table~\ref{tab:block_cap_results} shows the results of applying different block caps to our final model (SFT $\rightarrow$ DPO $\rightarrow$ GRPO).

\begin{table}[H]
    \begin{center}
        \begin{tabular}{@{}l|cc|c|ccc|cc@{}}
            \toprule
            \multirow{2}{*}{\textbf{Block Cap}} & \multicolumn{2}{c|}{\textbf{Length}} & \multirow{2}{*}{\textbf{Bad cases (\%)}} & \multicolumn{3}{c|}{\textbf{Accuracy}} & \multirow{2}{*}{\textbf{$\Delta$Acc}} & \multirow{2}{*}{\textbf{$\Delta$Len}}                                                                        \\
                                                & mean                                 & std                                      &                                        & Overall                               & Easy                                  & Difficult &                            &                             \\
            \midrule
            None (auto)                         & 5791                                 & 3160                                     & 3.8\%                                  & 85.3\%                                & 92.0\%                                & 74.9\%    & --                         & --                          \\
            $\leq 0$                            & 1046                                 & 1365                                     & 1.0\%                                  & 59.0\%                                & 52.0\%                                & 57.7\%    & \colorbox{red!15}{-26.3\%} & \colorbox{blue!15}{-81.9\%} \\
            $\leq 2$                            & 4860                                 & 2336                                     & 6.1\%                                  & 77.9\%                                & 73.6\%                                & 68.2\%    & -7.4\%                     & -16.1\%                     \\
            $\leq 6$                            & 5466                                 & 2849                                     & 3.3\%                                  & 80.4\%                                & 91.6\%                                & 74.0\%    & \colorbox{blue!15}{-4.9\%} & -5.6\%                      \\
            $ >6$                               & 7618                                 & 3711                                     & 3.5\%                                  & 80.2\%                                & 74.2\%                                & 77.0\%    & -5.1\%                     & \colorbox{red!15}{+31.5\%}  \\
            \bottomrule
        \end{tabular}
        \caption{Effect of inference-time block caps on the DeepMath~(1k) test set. A \emph{block cap} controls the number of reasoning blocks generated per solution: None (auto) lets the model decide; $\leq k$ limits the maximum to $k$ blocks; and $>6$ enforces at least seven blocks (deep-think). Columns report mean answer length (computed over correct predictions) and its standard deviation, the bad-case ratio, and accuracy on the overall/easy/difficult splits. $\Delta$Acc and $\Delta$Len denote changes relative to the None (auto) setting. The \emph{easy} split corresponds to difficulty levels 2.0--4.5 and the \emph{difficult} split to 8.0--9.0.}\label{tab:block_cap_results}
    \end{center}
\end{table}

The None (auto) setting is the default behavior of our final model, which automatically determines the number of blocks based on the problem difficulty. The $\leq 0$ setting forces no-think responses, which results in a significant drop in accuracy and an increase in bad cases. The $\leq 2$ setting reduces the mean length by 16.1\% but also decreases accuracy by 7.4\%, indicating that limiting reasoning depth can hurt performance. The $\leq 6$ setting achieves a good balance, reducing length by 5.6\% while maintaining a reasonable accuracy drop of 4.9\%. The $>6$ setting enforces deep-think responses, leading to longer answers (+31.5\%) but with a slight accuracy drop (-5.1\%). This setting particularly degrades performance on easy problems, which is expected since deep-think is rarely needed. In these failure cases, the model sometimes generates the correct answer early on, but the enforced long chain-of-thought can lead it to an incorrect final answer.

\section{Limitations}
\label{sec:limitations}

A key challenge is ensuring that the predicted block count consistently matches the actual number of reasoning blocks generated. This discrepancy arises from the inherent limitations of the LLM's instruction-following capabilities. Although we penalize mismatches during reinforcement learning to enforce format consistency, perfect adherence is not guaranteed. Future work will focus on developing more robust methods to align predicted and actual block counts, potentially through enhanced training techniques or architectural modifications.

\section{Conclusions}
\label{sec:conclusions}

Large language models often expend excessive computational resources on simple problems, a phenomenon known as "overthinking." To address this, we introduce a novel block-structured reasoning framework that enables LLMs to dynamically adapt their reasoning depth to match the difficulty of the problem at hand. By partitioning the reasoning process into discrete, controllable blocks, our model learns to predict the required number of reasoning steps and generate a solution accordingly.

Our three-stage training pipeline, consisting of supervised fine-tuning (SFT), Direct Preference Optimization (DPO), and Reinforcement Learning (RL), successfully teaches the model this adaptive behavior. Experimental results demonstrate the effectiveness of our approach. Our final model reduces the average answer length by a substantial 25.1\% while maintaining nearly the same level of accuracy as the baseline, with only a 0.2\% drop. This shows that the model can generate more concise answers for simpler problems without sacrificing performance on complex ones.

Furthermore, the explicit block structure provides a practical mechanism for inference-time control. Users can easily adjust the number of reasoning blocks to balance computational efficiency and solution accuracy, making the model more flexible and adaptable to different use cases. Our analysis confirms that the model naturally uses fewer blocks for easy problems and more for difficult ones, validating the effectiveness of our block-structured approach. This work represents a significant step towards more efficient and adaptive reasoning in large language models.

\bibliographystyle{plain}
\bibliography{refs}
\appendix
\section{Appendix}
\label{sec:appendix}

This appendix provides a detailed description.

\subsection{Prompt Engineering}

The prompt we used for the block-structured reasoning task is as follows:

\begin{promptbox}
    \textbf{\# Instruction: Segment Reasoning}

    \vspace{1em}

    \textbf{**Objective:**} \\
    Extract the reasoning process from the input \texttt{solution} field and divide it into several logically distinct segments based on major problem-solving steps.

    \hrulefill

    \textbf{**Input Data Structure:**} \\
    Each input item contains a \texttt{solution}. The \texttt{solution} field typically looks like:
    \begin{quote}
        \textit{Original, complete reasoning process text...}
    \end{quote}

    \textbf{**Output Format Specification:**} \\
    The processed \texttt{solution} field must be \textbf{structured} as follows (Reasoning segment i is part of \texttt{solution}):
    \begin{quote}
        Reasoning segment 1 (part of \texttt{solution}) \\
        \texttt{\textless continue\_think\textgreater} \\
        Reasoning segment 2 (part of \texttt{solution}) \\
        \texttt{\textless continue\_think\textgreater} \\
        … \\
        \texttt{\textless continue\_think\textgreater} \\
        Reasoning segment N (part of \texttt{solution})
    \end{quote}

    \hrulefill

    \textbf{**Detailed Requirements**}
    \begin{enumerate}
        \item \textbf{Segment by Problem-Solving Steps} \\
              Split the reasoning into distinct segments based on major problem-solving phases or logical steps in the solution process. Place \texttt{\textless continue\_think\textgreater} between segments. Focus on \textbf{substantial shifts in problem-solving approach} rather than minor semantic transitions. Examples of major logical boundaries include:
              \begin{itemize}
                  \item Moving from problem understanding to solution planning
                  \item Transitioning between different solution approaches or methods
                  \item Shifting from calculation to verification
                  \item Moving from one major sub-problem to another
                  \item Changing from analysis to synthesis
              \end{itemize}

        \item \textbf{Maximize Tokens per Segment}
              \begin{itemize}
                  \item Within the bounds of logical coherence, pack as much reasoning as possible into each segment.
                  \item Create \textit{fewer, longer} segments that capture complete problem-solving phases rather than fragmenting at every minor transition.
                  \item Insert \texttt{\textless continue\_think\textgreater} only when a major problem-solving step concludes or when beginning a fundamentally different approach to the problem.
              \end{itemize}

        \item \textbf{Segment Number Guidelines}: Insert at least \texttt{difficulty / 2} \texttt{\textless continue\_think\textgreater} separators; do not produce fewer than this minimum, but you are encouraged to insert more but no more than \texttt{2*difficulty}.

        \item \textbf{Preserve Content \& Ensure Coherence}
              \begin{itemize}
                  \item Do \textbf{not} add, delete, or modify any words in the original reasoning text—only insert \texttt{\textless continue\_think\textgreater} separators.
              \end{itemize}
    \end{enumerate}

    \hrulefill

    \textbf{**Task:**} \\
    Apply the above rules to the provided \texttt{solution} content. Output only the processed result—do not include any additional commentary, explanations, or extraneous text. Do \textbf{not} add, delete, or modify any words in the original reasoning text—only insert \texttt{\textless continue\_think\textgreater} separators.
\end{promptbox}

\end{document}